# IM-GIV: an effective integrity monitoring scheme for tightly-coupled GNSS/INS/Vision integration based on factor graph optimization

Yunong Tian, Tuan Li, Haitao Jiang, Zhipeng Wang, and Chuang Shi

*Abstract*—Global Navigation Satellite System/Inertial Navigation System (GNSS/INS)/Vision integration based on factor graph optimization (FGO) has recently attracted extensive attention in navigation and robotics community. Integrity monitoring (IM) capability is required when FGO-based integrated navigation system is used for safety-critical applications. However, traditional researches on IM of integrated navigation system are mostly based on Kalman filter. It is urgent to develop effective IM scheme for FGO-based GNSS/INS/Vision integration. In this contribution, the position error bounding formula to ensure the integrity of the GNSS/INS/Vision integration based on FGO is designed and validated for the first time. It can be calculated by the linearized equations from the residuals of GNSS pseudo-range, IMU pre-integration and visual measurements. The specific position error bounding is given in the case of GNSS, INS and visual measurement faults. Field experiments were conducted to evaluate and validate the performance of the proposed position error bounding. Experimental results demonstrate that the proposed position error bounding for the GNSS/INS/Vision integration based on FGO can correctly fit the position error against different fault modes, and the availability of integrity in six fault modes is 100% after correct and timely fault exclusion.

## I. INTRODUCTION

PRECISE navigation and its integrity is crucial for many applications such as intelligent vehicles, autonomous driving (AD) and unmanned aerial vehicles. Global Navigation Satellite System (GNSS) is well-known for providing globally referenced position. However, GNSS signals are easy to be interrupted or even totally unavailable. In order to improve GNSS performance in challenging environments, GNSS is generally integrated with Inertial Measurement Unit (IMU) due to their complementary characteristics[1-4]. In order to further improve the performance of GNSS/IMU integration in GNSS-challenged environments, the integration of GNSS, IMU, and vision has been a hot research topic in recent years[5-9]. Among the various approaches, the multi-sensor integration based on the FGO method has become predominant, favored for its higher performance merits.

The performance of the navigation systems can be quantified by the four properties: accuracy, availability, continuity and integrity, and integrity monitoring of the navigation system is closely related to its safety. Integrity monitoring consists of two parts: fault detection and the calculation of the protection level (PL). Fault detection provides the ability to alarm users when faults occur, and the faulty measurements are excluded to improve continuity [10]. To make PL easier to understand, three IM parameters are introduced below:

- **Alert limit (AL)**: The maximum allowable position error. The navigation system is unavailable when PL exceeds it.
- **Hazardously Misleading Information (HMI)**: Position error is larger than the AL while the PL is smaller than the AL.
- **Integrity risk (IR)**: The probability of HMI within time to alert.

Then, PL is the position error bounding, which is the critical value of the Probability Density Function (PDF) of the position error at a confidence interval of 1- IR [11].

The integrity monitoring of GNSS is divided into system level and user level [12-14]. Compared with the former, the latter does not use external systems or infrastructure. It is worth noting that the detection method for GNSS faults in this paper belongs to the user level. The typical representative of the user level is Receiver Autonomous Integrity Monitoring (RAIM)[15]. For RAIM, according to fault detection based on the measurements and position solution, RAIM can be categorized into residual based (RB) RAIM and Multiple Hypothesis Solution Separation (MHSS) RAIM [16]. Compared with the former, the latter has a high computational cost, but it is more rigorous [17]. According to the GNSS measurements from current epoch and past measurements, snapshot schemes and filtering schemes are also derived. The Least Squares (LS) RAIM belongs to first schemes, in which

This study was supported in part by the National Key Research and Development Program of China (Grant No. 2023YFB3907302). (Corresponding author: Tuan Li).

Yunong Tian and Tuan Li are with the Advanced Research Institute of Multidisciplinary Sciences, Beijing Institute of Technology, Beijing 100081, China (e-mail: yunongtian@bit.edu.cn; tuanli@buaa.edu.cn).

Haitao Jiang, Zhipeng Wang, and Chuang Shi are with the School of Electronic and Information Engineering, Beihang University, Beijing 100191, China (e-mail: by1902019@buaa.edu.cn; wangzhipeng@buaa.edu.cn; shichuang@buaa.edu.cn).

the navigation solutions obtained by LS estimation were monitored [18, 19]. The PL consists of two parts, the influence of GNSS measurement noise and bias. It can effectively reflect position errors of the system. The second scheme is filtering based RAIM. The extended Kalman filtering (EKF) method is normally used to estimate the states of the navigation system [20, 21]. This scheme can obtain better PL estimations than LS RAIM scheme. However, the computation burden is large in some cases.

For the GNSS/Inertial Navigation System (INS) integration, early researches have focused on IMU-assisted RAIM to improve satellite fault detection rate and reduce GNSS IR under the assumption that the IMU measurements are not faulty[22]. The same as LS RAIM, the PL of integrated GNSS/INS system also includes two parts: the influence term of GNSS measurements noise and bias [23-25]. The former can be calculated by the EKF covariance matrix and the latter was represented by the upper bound of the measurement bias. In MHSS, the horizontal protection level (HPL) is affected by both the differences between subset solutions and each solution's covariance [23]. However, this method is complex, and the assumption of position error that calculated by the difference of the full set solution and the subset solution being a Gaussian variable has not been resolved fully yet in some case. Moreover, for low-dimensional systems, it has no advantage compared with other methods. With the development of the AD, and the possible IMU fault risk[26, 27], the integrity impacts of IMU faults were considered in recent years. To assure integrity of UAV navigation systems, Liu et al. [28] proposed a real-time algorithm to calculate the vertical protection level based on EKF against IMU sensor faults. The real time vertical protection level (VPL) can over-bound the true vertical position errors during UAV operations. Gao et al. [29] proposed a new IM method to deal with the problem that the assumption that measurement noise is the Gaussian white noise is not true. However, the availability of integrity is low due to the value of the PL is tens of meters. For the integrity of GNSS/INS/Vision integration, few works have been conducted. In [30], we proposed an effective IM method for the EKF-based GNSS/INS/Vision system, and the results show that the proposed algorithm is effective against six different fault modes [31].

As stated above, GNSS/INS/Vision integration via optimization method has attracted extensive attention due to its high accuracy and robustness. However, the corresponding integrity monitoring scheme has not been explored. In this contribution, an effective integrity monitoring scheme is proposed for FGO-based GNSS/INS/Vision integration. The main contributions of this paper are summarized as follows:

(1) The IM scheme for the tightly-coupled GNSS/INS/Vision integration based on FGO is presented.

(2) A new method to calculate the position error bounding of the FGO-based GNSS/INS/Vision integration against sensor faults is proposed for the first time.

(3) Extensive field experiments were conducted to validate the proposed method.

We first present the tightly-coupled GNSS/INS/Vision system model and describe the IMU, visual and pseudo-range measurements and its residuals. Moreover, we describe the fault detection method and the classification process of fault mode in detail. Then, we derive the position error bounding formulas for different fault modes exhaustively. Next, we present the architecture of the system with IM, including fault detection and the calculation of position error bounding. A detailed analysis and evaluation of position error bounding over multiple-sets data completes this section. Finally, conclusions are drawn.

II. FGO-BASED GNSS/INS/VISION INTEGRATION MODEL

In this section, to make the follow notations clear, we first introduce the coordinate frames [6] used in this research. It is shown in Fig. 1.

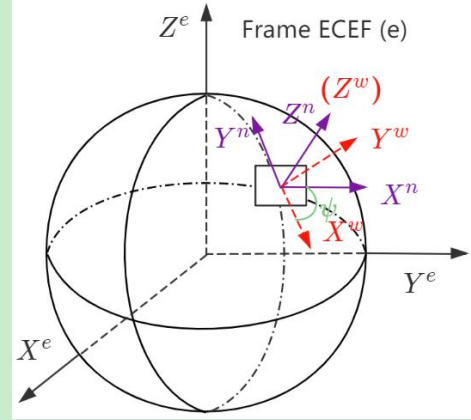

**Fig. 1.** Diagram of coordinate frames. (Earth-Centered Earth-Fixed, e), (Local World Frame, $w$), (ENU Frame, n), $\psi$ is the yaw offset between $w$ frame and $n$ frame.

- **Sensor Frame:** It includes camera frame (c-frame) and IMU frame ($b$-frame). Besides, IMU frame at time $k$ is marked as frame $b_k$ and camera frame at time $k$ is marked as frame $c_k$.
- **Local World Frame:** The frame where visual-inertial system operates is marked as the local world frame ($w$-frame). As illustrated in Fig. 1, its origin is arbitrarily set and the z axis is often chosen to be orthogonal to the local ground plane.
- **ECEF Frame:** The origin of ECEF (Earth-Centered Earth-Fixed, e-frame) frame is attached to the center of mass of Earth. The x-y plane coincides with Earth's Equator, and the z axis points to the true north.
- **ENU Frame:** The ENU (East-North-Up, $n$-frame) frame is used to connect $w$ and $e$ frame. $\psi$ is the yaw offset between $w$ and n frame. The X-axis, Y-axis and Z-axis of $n$ frame points East, North and Up direction, respectively.

The IMU, visual, and pseudo-range measurements and their corresponding residuals are described in detail by Cao et al. [32]. The state vector of the system is represented as:

$$\begin{cases} \chi = [x_1 \cdots x_n \ \rho_1 \cdots \rho_n \ \psi] \\ x_k = [p_{b_k}^w \ v_{b_k}^w \ q_{b_k}^w \ b_a \ b_g \ \delta t \ \dot{\delta t}], \quad k\epsilon[1,n] \\ \delta t = [\delta t_G \ \delta t_R \ \delta t_E \ \delta t_C] \end{cases} \quad (1)$$

where $x_k$ denotes the IMU state at the time that the k-th image is captured; $p_{b_k}^w$, $v_{b_k}^w$ and $q_{b_k}^w$ are the position, velocity and orientation of the IMU in the $w$ frame; $b_a$ and $b_g$ are the accelerometer, gyroscope bias term of the IMU; $\delta t$ and $\dot{\delta t}$ denote the receiver clock bias and receiver clock drifting rate, respectively; $\rho_m$ stands for the inverse distance of the m-th feature; $\delta t_G$, $\delta t_R$, $\delta t_E$ and $\delta t_C$ denote the clock bias from GPS, GLONASS, GALILEO and BeiDou systems, respectively. The subscript n represents the window size; the subscript m denotes the total number of feature points.

*A. IMU Pre-integration Measurement Model*

Since the IMU, Visual, and GNSS measurements are fused using FGO-based method, the observation residuals are formulated to estimate the navigation solution. The IMU measurements in two consecutive frames are $Fr_k$ and $Fr_{k+1}$; thus, the residual of the IMU pre-integration measurement can be written as [33]:

$$r_B\left(\hat{z}_{Fr_{k+1}}^{Fr_k}, \chi\right) = \begin{bmatrix} \delta\alpha_{Fr_{k+1}}^{Fr_k} \\ \delta\beta_{Fr_{k+1}}^{Fr_k} \\ \delta\theta_{Fr_{k+1}}^{Fr_k} \\ \delta b_a \\ \delta b_g \end{bmatrix} \quad (2)$$

where $\hat{z}_{Fr_{k+1}}^{Fr_k}$ denotes pre-integrated IMU measurements; $\delta\alpha_{Fr_{k+1}}^{Fr_k}$, $\delta\beta_{Fr_{k+1}}^{Fr_k}$, and $\delta\theta_{Fr_{k+1}}^{Fr_k}$ represent the residual of pre-integrated IMU measurements between two consecutive frames; $\delta b_a$ and $\delta b_g$ denote accelerometer and gyroscope biases, respectively.

*B. Visual Measurement Model*

If a feature $l$ observed in image frame $c_i$, it is observed again in image frame $c_j$, the projection process can be modelled as:

$$\widetilde{\mathcal{P}}_l^{c_j} = \pi_c(\ \mathbf{R}_b^c(\mathbf{R}_w^b x^w + t_w^b) + t_b^c\ ) + n_c \quad (3)$$

where $\widetilde{\mathcal{P}}_l^{c_j}$ denotes the feature coordinate in the j-th image plane and $x^w$ denotes its position in the $w$ frame; $\pi_c(\cdot)$ represents the camera projection function; $n_c$ stands for the measurement noise; $\{\mathbf{R}_w^b, t_w^b\}$ and $\{\mathbf{R}_b^c, t_b^c\}$ represent the transformation from the $w$ frame to the body coordinate frame and from the body coordinate frame to the camera coordinate frame, respectively;

The residual of the feature measurement in the j-th image can be written as:

$$r_C(\tilde{z}_l, \chi) = \widetilde{\mathcal{P}}_l^{c_j} - \pi_c(\hat{x}_l^{c_j}) \quad (4)$$

where $\tilde{z}_l$ denotes visual measurements, and it is related to two frames; $\hat{x}_l^{c_j}$ denotes the estimated value of the feature in frame $j$; it is relevant to the measurement of the feature in the i-th image.

*C. Pseudo-Range Measurements*

The pseudo-range (PR) measurements between the satellite $s_j$ and the GNSS receiver in the earth-centered earth-fixed frame is defined as [6]:

$$\begin{aligned}\tilde{p}_{r_k}^{s_j} = \ & \|p_{s_j}^e - p_{r_k}^e\| + l\_s \cdot (\delta t - \Delta t^{s_j}) \\ & + T_{r_k}^{s_j} + I_{r_k}^{s_j} + M_{r_k}^{s_j} + S_{r_k}^{s_j} + \epsilon_{r_k}^{s_j}\end{aligned} \quad (5)$$

where $p_{s_j}^e$ and $p_{r_k}^e$ represent the position of the satellite $s_j$ and receiver $r$, respectively; $l\_s$ denotes the speed of light; $\Delta t^{s_j}$ denotes the satellite's clock offset; $T_{r_k}^{s_j}$ and $I_{r_k}^{s_j}$ represent the tropospheric and ionospheric delay, respectively; $M_{r_k}^{s_j}$ and $S_{r_k}^{s_j}$ denote the multipath delay and Sagnac effect, respectively; $\epsilon_{r_k}^{s_j}$ represents measurement noise, and it is assumed to be Gaussian distributed with zero-mean and covariance $\sigma_{r_k,pr}^{s_j}$.

The residual of the pseudo-range measurement at epoch k is defined as:

$$\begin{aligned}r_p\left(\tilde{z}_{r_k}^{s_j}, \chi\right) = \ & \|p_{s_j}^e - \mathbf{R}_n^e \mathbf{R}_w^n p_{b_k}^w - p_{anc}^e\| + \\ & l\_s \cdot (\delta t - \Delta t^{s_j}) + T_{r_k}^{s_j} + I_{r_k}^{s_j} + S_{r_k}^{s_j} - \tilde{p}_{r_k}^{s_j}\end{aligned} \quad (6)$$

where $\tilde{z}_{r_k}^{s_j}$ represents the pseudo-range measurement of the satellite $s_j$; $n$ denotes the ENU frame; $\mathbf{R}_n^e$ and $\mathbf{R}_w^n$ denote the transformation matrix from the $n$ to the ECEF frame and from $w$ frame to the $n$ frame, respectively; $p_{anc}^e$ denotes the position of the anchor point in e frame.

III. INTEGRITY MONITORING SCHEME FOR FGO-BASED GNSS/INS/VISION INTEGRATION

*A. Fault Detection Method*

For GNSS, we only consider the faults in the pseudorange measurements because the doppler measurements are less sensitive to multipath effects and non-line of sight (NLOS) receptions. Then, the normalized GNSS pseudorange measurement residual vector inside a sliding window with a size of M epochs can be denoted as [34]:

$$e_{M,p} = \left(\hat{e}_{p,1}^T \ \ldots \ \hat{e}_{p,k}^T\right)^T \quad (7)$$

$\hat{e}_{p,k}$ can be expressed as:

$$\hat{e}_{p,k} = \left(\hat{e}_{p,k}^{s_1} \ \ldots \ \hat{e}_{p,k}^{s_j}\right)^T \quad (8)$$

where $\hat{e}_{p,k}^{s_j}$ denotes normalized pseudorange measurement residual vector of $s_j$-th satellite in epoch k.

The test statistics of $e_{M,p}$ is calculated as:

$$T_{S,M}^p = e_{M,p}^T \cdot e_{M,p} \quad (9)$$

According to the above formula, the normalized IMU pre-integration and visual measurement residual vector in a sliding window with a size of $M$ epochs are written as:

$$e_{M,IMU} = (\hat{e}_{IMU,1} \ \ldots \ \hat{e}_{IMU,k})^T \quad (10)$$
$$e_{M,cam} = (\hat{e}_{cam,1} \ \ldots \ \hat{e}_{cam,k})^T \quad (11)$$

where $\hat{e}_{IMU,k}$ denotes normalized IMU pre-integration measurements residual vector in epoch $k$; $\hat{e}_{cam,k}$ denotes normalized visual measurements residual vector in epoch k.

The test statistics of $e_{M,IMU}$ and $e_{M,cam}$ are calculated as:

$$T_{S,M}^{IMU} = e_{M,IMU}^T \cdot e_{M,IMU} \quad (12)$$
$$T_{S,M}^{cam} = e_{M,cam}^T \cdot e_{M,cam} \quad (13)$$

It is worth noting that the false alarm probability ($P_{FA}$) and the miss detection probability ($P_{MD}$) is set to $10^{-5}/h$ and $10^{-3}/h$[35], and the detection threshold is determined by the false alarm probability and the degree of freedom of the corresponding residual vector. These test statistics will be compared with the corresponding detection threshold to determine whether the corresponding measurements are faulty or not. Additionally, the faults will be excluded immediately once they are detected.

*B. Fault mode classification and Integrity Risk Allocation*

For the GNSS/INS/Vision integration, there are eight fault modes:
$$F_i = \{GIV\ \overline{G}IV\ G\overline{I}V\ GI\overline{V}\ G\overline{I}\overline{V}\ \overline{G}\overline{I}V\ \overline{G}I\overline{V}\ \overline{G}\overline{I}\overline{V}\} \quad (14)$$
where $G$ and $\overline{G}$ denote satellite measurements with and without faults, respectively; $I$ and $\overline{I}$ denote IMU measurements with and without faults, respectively; $V$ and $\overline{V}$ denote visual measurements with and without faults, respectively. The total integrity risk of this system can be expressed as the sum of the integrity risks under different fault modes:
$$P(HMI) = \sum_{i=1}^{8} P(HMI, F_i) = \sum_{i=1}^{8} P(HMI \mid F_i) P(F_i) \quad (15)$$
where $P(F_i)$ is the prior probability of $F_i$; $P(HMI \mid F_i)$ is the probability that HMI will occur in the case of $F_i$.
Take the horizontal integrity risk as an example, $P(HMI \mid F_i)$ can be written as:
$$P(HMI \mid F_i) = P(|HPE| \geq HAL | F_i) \quad (16)$$
where $HPE$ and $HAL$ are the horizontal positioning error and the horizontal alarm limit, respectively.

In this research, the total integrity risk $P_{HMI,Total}$ of the system is set to $1 \times 10^{-7}/h$[36], which is allocated 50% to the vertical domain ($P_{HMI,V}$) and 50% to the horizontal domain ($P_{HMI,H}$). Considering the hardware qualities used for navigation system [37, 38], the prior probabilities of single satellite, IMU and visual measurement with fault were assumed to be $P(G) = 10^{-5}$, $P(I) = 10^{-3}$ and $P(V) = 10^{-4}$, respectively. It is worth noting that the total integrity risk does not need to be allocated to all fault modes due to the fact that the probability of some fault modes is very low.

Then, taking $\overline{G}\overline{I}\overline{V}$ and $G\overline{I}\overline{V}$ fault modes as examples, we introduce the process that whether the corresponding fault mode needs to be allocated integrity risk.

For $\overline{G}\overline{I}\overline{V}$ fault mode, $P(HMI, \overline{G}\overline{I}\overline{V})$ can be expressed as:
$$P(HMI, \overline{G}\overline{I}\overline{V}) = P(HMI \mid \overline{G}\overline{I}\overline{V}) * P(\overline{G}\overline{I}\overline{V}) \quad (17)$$
Assuming that satellite, IMU and visual measurement faults are independent from each other in the current epoch, $P(\overline{G}\overline{I}\overline{V})$ can be written as:
$$P(\overline{G}\overline{I}\overline{V}) = P(\overline{G}) * P(\overline{I}) * P(\overline{V}) \quad (18)$$
where $P(\overline{G}) = 1 - P(G)$, $P(\overline{I}) = 1 - P(I)$ and $P(\overline{V}) = 1 - P(V)$ denotes the probability of single satellite, IMU and vision measurement without faults. $P(HMI, \overline{G}\overline{I}\overline{V})$ can be rewritten as:
$$P(HMI, \overline{G}\overline{I}\overline{V}) = 0.9989 * P(HMI \mid \overline{G}\overline{I}\overline{V}) \quad (19)$$
From (19), we know that $P(HMI, \overline{G}\overline{I}\overline{V})$ may be larger than the total integrity risk. Therefore, it needs to be allocated integrity risk.

For $G\overline{I}\overline{V}$ fault mode, $P(HMI, G\overline{I}\overline{V})$ can be expressed as:

$$P(HMI, G\overline{I}\overline{V}) = 0.999 * 10^{-9} * P(HMI \mid G\overline{I}\overline{V}) \quad (20)$$
The result shows that $P(HMI, G\overline{I}\overline{V})$ is much smaller than the total integrity risk. Therefore, it not need to be allocated integrity risk. By repeating the above process, we find that the integrity risk of the system needs to be allocated to the following fault modes: $\overline{G}\overline{I}\overline{V}$, $\overline{G}\overline{I}V$, $G\overline{I}\overline{V}$, $\overline{G}I\overline{V}$, $\overline{G}IV$, $\overline{G}I\overline{V}$. It is worth noting that the horizontal integrity risk is equally allocated to the above six fault modes.

*C. Position Error Bounding*

In order to evaluate the integrity risk of the FGO-based GNSS/INS/Vision integration, the protection level formula for six different fault modes is derived in this section.

(1) $\overline{G}IV$ fault mode

In the $\overline{G}IV$ fault mode, the linearized equation affected by the IMU pre-integration measurement residual in epoch k is expressed as:
$$J_{I,k} \cdot \delta x_k = e_{i,k} \quad (21)$$
$$\delta x_k = (J_{I,k}^T W_I J_{I,k})^{-1} J_{I,k}^T W_I e_{I,k} \quad (22)$$
where $I$ denotes IMU sensor; $J_{I,k}$ denotes Jacobian matrix of $r_B\left(\hat{z}_{Fr_{k+1}}^{Fr_k}, \chi\right)$ relating to state $x_k$; $e_{I,k}$ denotes the difference between the IMU pre-integration measurements and the predicted values; $W_I$ denotes the inverse of the covariance matrix of measurements noise.

In order to simplify (22), the matrix $\mathbf{B}_{I,k}$ is defined as:
$$\mathbf{B}_{I,k} = (J_{I,k}^T W_I J_{I,k})^{-1} J_{I,k}^T W_I \quad (23)$$
Then, (22) can be rewritten as:
$$\delta x_k = \mathbf{B}_{I,k} e_{I,k} \quad (24)$$
Assuming that the state errors $\varepsilon_{i,k}$ are caused by IMU pre-integration measurement noise $\epsilon_{i,k}$, (21) can be rewritten as:
$$J_{I,k} \cdot (\delta x_k + \varepsilon_{I,k}) = e_{I,k} + \epsilon_{I,k} \quad (25)$$
Then, we obtain the following formular using (21) and (24):
$$\varepsilon_{I,k} = \mathbf{B}_{I,k} \epsilon_{I,k} \quad (26)$$
The covariance matrix of the state error $\varepsilon_{I,k}$ affected by $\epsilon_{I,k}$ is obtained as:

$$C_{r_B} = (J_{I,k}^T J_{I,k})^{-1} \sigma_I^2 \quad (27)$$
where $\sigma_I^2$ is the variance of $\epsilon_{I,k}$.

From (27), the covariance matrix of the state error affected by visual measurement noise $\epsilon_{c,k}$ and pseudo-range measurement noise $\epsilon_{p,k}$ can be written as:
$$\begin{cases} C_{r_C} = (J_{c,k}^T J_{c,k})^{-1} \sigma_c^2 \\ C_{r_p} = (J_{p,k}^T J_{p,k})^{-1} \sigma_p^2 \end{cases} \quad (28)$$
where $c$ denotes camera; $p$ denotes pseudo-range; $J_{p,k}$ denotes Jacobian matrix of $r_p\left(\tilde{z}_{r_k}^{S_j}, \chi\right)$ relating to state $[p_{b_k}^w\ \delta t_k\ \psi]$; $J_{c,k}$ represents Jacobian matrix of $r_C(\tilde{z}_l, \chi)$ with respect to state $p_{b_k}^w$; $\sigma_c^2$ and $\sigma_p^2$ represent the variance of visual and pseudo-range measurements, respectively.

Therefore, the final covariance matrix of the state error affected by $\epsilon_{I,k}$, $\epsilon_{c,k}$, and $\epsilon_{p,k}$ can be expressed as:
$$P = C_{r_B} + C_{r_C} + C_{r_p} \quad (29)$$
Then, the position error bounding can be expressed as [38]:

$$PEB_{q,\overline{GIV}} = \pm K_{md,\overline{GIV}} * \xi_q \qquad (30)$$

where $'q'$ denotes the q-th component of $\delta x_k$; $\xi_q$ is the square root of the q-th diagonal elements of $P$; $K_{md,\overline{GIV}}$ is the bilateral quantile of the normal distribution corresponding to $P(HMI \mid \overline{GIV})$.

(2) $\overline{G}\overline{I}\overline{V}$ fault mode

When the IMU pre-integration measurements have fault $f_{I,k}$ in epoch $k$, we know from (26) that the state errors affected by the IMU pre-integration measurement noise $\epsilon_{I,k}$ and $f_{I,k}$ can be written as:

$$\varepsilon_{I,k} + \varepsilon_{fI,k} = \mathbf{B}_{I,k}\epsilon_{I,k} + \mathbf{B}_{I,k}f_{I,k} \qquad (32)$$

Unfortunately, $f_{I,k}$ is unknown. $\mathbf{B}_{I,k}f_{I,k}$ can be obtained as the result of the maximum characteristic slope $(Slope_{mq,i})$ multiplied by the minimum measurable deviation $\sqrt{\lambda_{a,I}}$ [39]. From (32), the influence of the m-th element of $f_{I,k}$ on the $q$ dimension of the state error is:

$$\mu_q = \left(\mathbf{B}_{I,k}\right)_{qm}\xi_b = b_{qm,I}\xi_b \qquad (33)$$

where $\xi_b$ is the element in the $b$ dimension of $f_{I,k}$; $b_{qm,I}$ denotes the element in row q and column m of $\mathbf{B}_{I,k}$.

The test statistic $T_{s,I}$ affected by $\xi_b$ can be written as [39]:

$$\lambda = \xi_b^2 \tau_{mm,I} \qquad (34)$$

where $\tau_{mm,I}$ denotes the $m$ row and $m$ column element of the inverse matrix of the covariance matrix of $r_B\left(\hat{\mathbf{z}}_{Fr_{k+1}}^{Fr_k}, \chi\right)$.

The characteristic slope can be calculated by the ratio of the state estimation error caused by the IMU pre-integration measurement faults to $T_{s,I}$, and $\sqrt{\lambda_{a,I}}$ is related to the probability of missed detection. $Slope_{mq,I}$ is defined as:

$$Slope_{mq,I} = \sqrt{\frac{\mu_q^2}{\lambda}} = \frac{|b_{qm,I}|}{\sqrt{\tau_{mm,I}}} \qquad (35)$$

According to the above analysis, the position error bounding against the IMU pre-integration measurement faults can be rewritten as:

$$PEB_{q,\overline{G}I\overline{V}} = K_{md,\overline{G}I\overline{V}} * \xi_q + Max(Slope_{mq,I}) * \sqrt{\lambda_{a,I}} \qquad (36)$$

where $K_{md,\overline{G}I\overline{V}}$ is the bilateral quantile of the normal distribution corresponding to $P(HMI \mid \overline{G}I\overline{V})$.

(3) $G\overline{IV}$ fault mode

When the GNSS pseudo-range measurements have fault $f_{p,k}$ in epoch $k$, the state errors affected by $\epsilon_{p,k}$ and $f_{p,k}$ can be written from (32) as:

$$\varepsilon_{p,k} + \varepsilon_{fp,k} = \left(\mathbf{J}_{p,k}^T \mathbf{W}_p \mathbf{J}_{p,k}\right)^{-1} \mathbf{J}_{p,k}^T \mathbf{W}_p \cdot (\epsilon_{p,k} + f_{p,k}) \qquad (37)$$

where $\mathbf{W}_p$ denotes the inverse of the covariance matrix of $\epsilon_{p,k}$. Then, from (36), the corresponding position error bounding is expressed as:

$$PEB_{q,G\overline{IV}} = K_{md,G\overline{IV}} * \xi_q + Max(Slope_{mq,p}) * \sqrt{\lambda_{a,p}} \qquad (38)$$

where $K_{md,G\overline{IV}}$ is the bilateral quantile of the normal distribution corresponding to $P(HMI \mid G\overline{IV})$; $\sqrt{\lambda_{a,p}}$ is related to a given probability of MD of the pseudo-range measurement faults, and the definition of $Slope_{mq,p}$ is similar to $Slope_{mq,I}$.

(4) $\overline{GIV}$ fault mode

When the visual measurements have fault $f_{c,k}$ in epoch $k$, the state errors affected by $\epsilon_{c,k}$ and $f_{c,k}$ at time k can be written as:

$$\varepsilon_{c,k} + \varepsilon_{fc,k} = \left(\mathbf{J}_{c,k}^T \mathbf{W}_c \mathbf{J}_{c,k}\right)^{-1} \mathbf{J}_{c,k}^T \mathbf{W}_c \cdot (\epsilon_{c,k} + f_{c,k}) \qquad (39)$$

where $\mathbf{W}_p$ denotes the inverse of the covariance matrix of $\epsilon_{c,k}$. Then, the position error bounding can be written as:

$$PEB_{q,\overline{GIV}} = K_{md,\overline{GIV}} * \xi_q + Max(Slope_{mq,c}) * \sqrt{\lambda_{a,c}} \qquad (40)$$

where $K_{md,\overline{GIV}}$ is the bilateral quantile of the normal distribution corresponding to $P(HMI \mid \overline{GIV})$; $\sqrt{\lambda_{a,c}}$ can be calculated by a given probability of MD of the visual measurement faults, and the definition of $Slope_{mq,c}$ is also similar to $Slope_{mq,I}$.

(5) $GI\overline{V}$ and $\overline{G}IV$ fault mode

Base on the above analysis, the position error bounding formula for the $GI\overline{V}$ and $\overline{G}IV$ fault mode can be derived as:

$$PEB_{q,GI\overline{V}} = K_{md,GI\overline{V}} * \xi_q +$$
$$Max(Slope_{mq,i}) * \sqrt{\lambda_{a,i}} + Max(Slope_{mq,p}) * \sqrt{\lambda_{a,p}} \qquad (41)$$

$$PEB_{q,\overline{G}IV} = K_{md,\overline{G}IV} * \xi_q +$$
$$Max(Slope_{mq,i}) * \sqrt{\lambda_{a,i}} + Max(Slope_{mq,c}) * \sqrt{\lambda_{a,c}} \qquad (42)$$

where $K_{md,GI\overline{V}}$ and $K_{md,\overline{G}IV}$ are the bilateral quantile of the normal distribution corresponding to $P(HMI \mid GI\overline{V})$ and $P(HMI \mid \overline{G}IV)$.

## IV. OVERVIEW OF THE GNSS/INS/VISION INTEGRATION SYSTEM WITH INTEGRITY MONITORING

To understand the system clearly, Fig. 2 shows the structure of the FGO-based GNSS/INS/Vision integration with integrity monitoring.

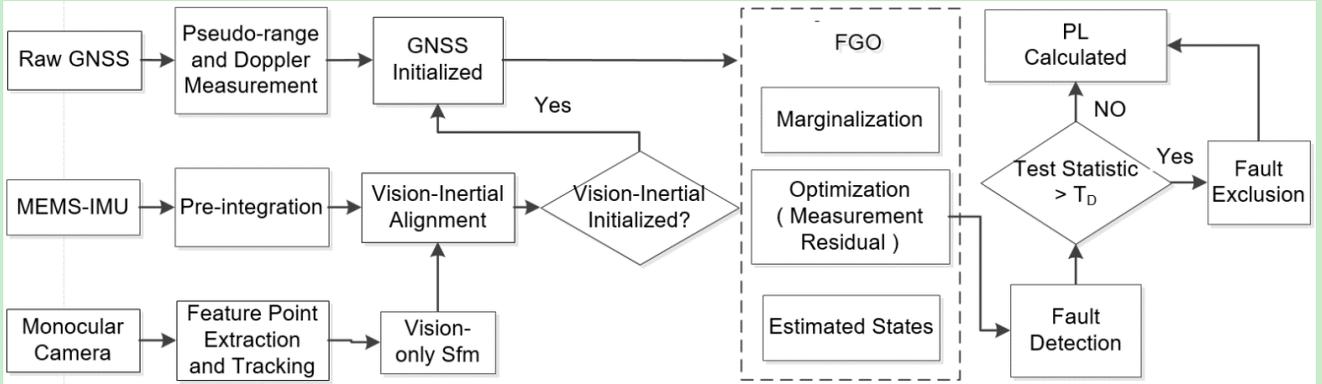

**Fig. 2.** Block diagram illustrating of the tightly-coupled GNSS/INS/Vision integration with integrity monitoring. PEB (Position Error Bounding), GNSS (Global Navigation Satellite System), MEMS-IMU (Micro-Electro-Mechanical-System Inertial Measurement Unit), TD (Test Statistic Threshold, FGO (Factor graph optimization)).

The system mainly involves four parts: the first part is data input process, including raw GNSS, IMU, and visual measurements, and the necessary preprocessing on each type of measurement is performed afterwards. The second part is vision-inertial initialization. After vision-inertial initialization is finished, a coarse position result can be obtained by the Single Point Positioning (SPP) algorithm. The third part is FGO-based state estimation. Constraints from the measurements in the sliding window are optimized under the non-linear optimization framework [32]. The fourth part is the IM of the GIV system. The system's integrity monitoring includes fault detection and calculation of the position error bounding. The corresponding test statistic can be calculated by the residual of pseudo-range, IMU pre-integration and visual measurement [34]. Then, test statistic is compared with predefined threshold to determine whether the measurements are faulty. The faults are detected and the corresponding measurements are excluded in time. Finally, the position error bounding is calculated to assure the integrity of the GIV system.

## V. FIELD TEST AND EXPERIMENTAL RESULT

We validate the performance of the propose IM method based on the public GVINS dataset "sports field" [6]. Fig. 3 displays the number of satellites of the four main satellite navigation systems (GPS, GLONASS, Galileo and BeiDou) and the Position Dilution of Precision (PDOP) in the experiment. The PDOP values are small, and many satellites are available, indicating favorable conditions for GNSS positioning.

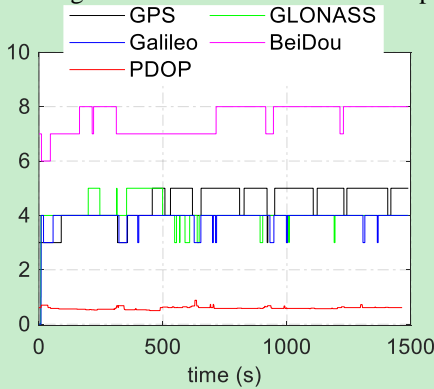

**Fig. 3.** Number of satellites of the four main satellite navigation systems (GPS, GLONASS, Galileo and BeiDou) and the Position Dilution of Precision (PDOP) in the experiment.

To verify the availability of integrity under different sensor faults, we present and analyze the position error bounding for the $\overline{GIV}$, $G\overline{IV}$, $GI\overline{V}$, $\overline{GIV}$, $\bar{G}IV$ and $\bar{G}I\bar{V}$ fault modes.

$\overline{GIV}$ *fault mode*

In the $\overline{GIV}$ fault mode, there are no sensor faults sensor measurements. Position error and the corresponding position error bounding in the horizontal direction are shown in Fig. 4.

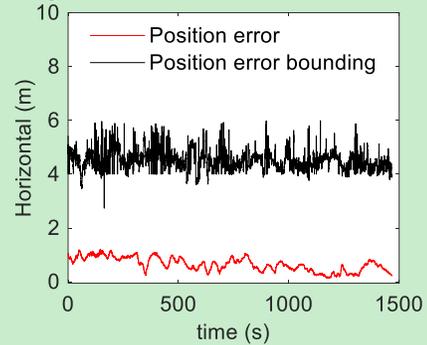

**Fig. 4.** Position error and the position error bounding in the horizontal direction when there are no sensor faults

Fig. 4 shows that the horizontal position error is enveloped by the corresponding position error bounding. The position error bounding in (30) considering the influence of the noise of the GNSS, visual and IMU pre-integration measurement on the position error. The result shows that the position error bounding assures the system's integrity when there are no faults in sensor measurements.

$\bar{G}I\bar{V}$ *fault mode*

In this fault mode, we manually add faults to the IMU measurements. A step fault with amplitude of $0.15~m/s^2$ for the accelerometer and 0.02 rad/s for the gyroscope is added to all three axes from 270 s to 290 s and 1110 s to 1130 s, respectively. Fig. 5 presents the horizontal position error and the corresponding position error bounding after fault detection and exclusion (FDE).

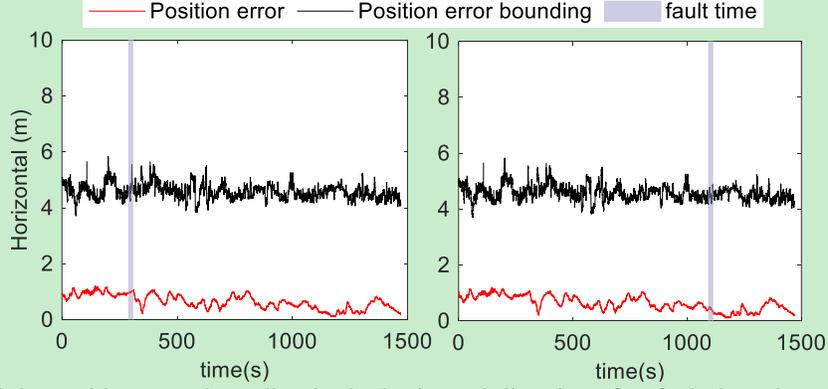

**Fig. 5.** Position error and the position error bounding in the horizontal direction after fault detection and exclusion. A step fault with amplitude of 0.15 $m/s^2$ for the accelerometer and 0.02 rad/s for the gyroscope is added to all three axes from 270 s to 290 s (left) and 1110 s to 1130 s (right)

Fig. 5 shows that the proposed position error bounding protects against the position error in the horizontal direction when step faults with amplitude of 0.15 $m/s^2$ for the raw accelerometer measurements and 0.02 rad/s for the gyroscope measurements at different times. This is due to the fact that faults are detected and excluded immediately, and the position error bounding obtained from (36) is related to the covariance matrix of the position error.

To further verify the effectiveness of the proposed method, Fig. 6 presents the horizontal position error and the corresponding position error bounding before FDE.

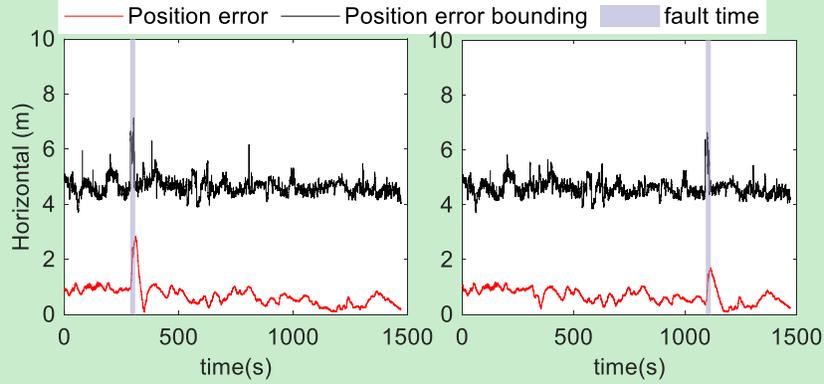

**Fig. 6.** Position error and the position error bounding in the horizontal direction before fault detection and exclusion. A step fault with amplitude of 0.15 $m/s^2$ for the accelerometer and 0.02 rad/s for the gyroscope is added to all three axes from 270 s to 290 s (left) and 1110 s to 1130 s (right)

Fig. 6 shows that the position error of the system in fault time increases significantly when IMU faults are not excluded. Also, the position error bounding in this case envelops the position error. The reason is that the position error bounding obtained from (36) is not only related to the covariance matrix of the position error but also the effect of IMU measurement bias on the position error. The results indicate that the proposed position error bounding protects against IMU measurement faults in the tightly-coupled GNSS/INS/Vision integration based on FGO.

*$G\overline{IV}$ fault mode*

As faulty satellites from different GNSS constellations have negligible influence on the position error, we take faulty measurements from GPS as an example. A step fault with amplitude of 15 m is added manually to all the pseudorange measurements on a single satellite from 380 s to 400 s and 1080 s to 1100 s. Fig. 7 shows the position error and the position error bounding in the horizontal direction after FDE.

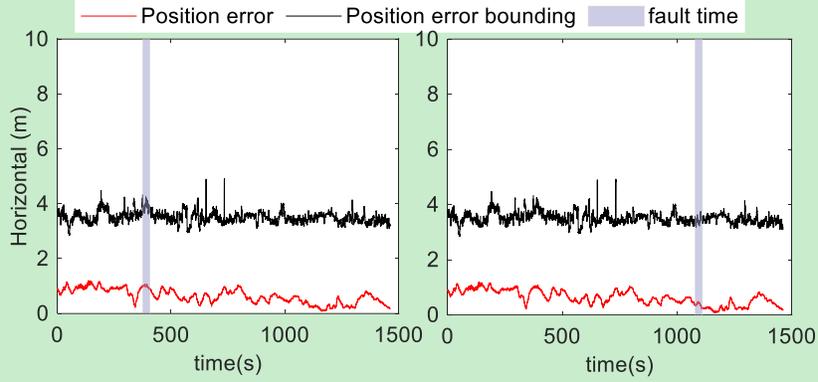

**Fig. 7.** Position error and the position error bounding in the horizontal direction after fault detection and exclusion. Step fault with amplitude of 15 m from 380 s to 400 s (left) and 1080 s to 1100 s (right) are added to pseudo-range measurements of a single satellite

Fig. 7 indicates that the position error bounding envelopes the corresponding position error after FDE. Moreover, in contrast to the $\overline{GIV}$ fault mode, there is no difference in the position error when all the pseudo-range measurements of a single satellite are faulty. The reason is that the pseudo-range measurements of a single satellite are immediately excluded when faults are detected.

Fig. 8 presents the position error and the corresponding position error bounding in the horizontal direction before FDE.

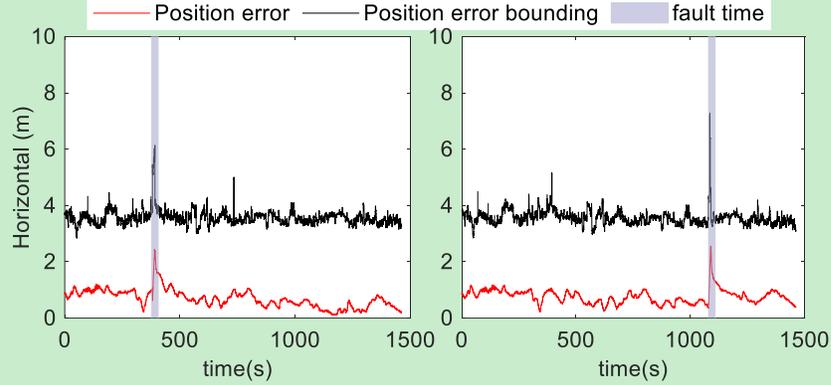

**Fig. 8.** Position error and the position error bounding in the horizontal direction before fault detection and exclusion. A step fault with amplitude of 15 m from 380 s to 400 s (left) and 1080 s to 1100 s (right) is added to all the pseudo-range measurements of a single satellite

From Fig. 8, we can see that as the fault time increases, the system's position error increases when faulty pseudo-range measurements are not excluded. However, the position error bounding always envelops the position error because of the influence of the second term in (38). Overall, the proposed position error bounding protects against the faulty pseudo-range measurements of a single satellite in the GNSS/INS/Vision integration.

*$GI\overline{V}$ fault mode*

In this fault mode, the above IMU fault and GNSS fault are added to the IMU measurements and single satellite pseudorange measurements from 700 s to 720 s, respectively. Fig. 9 shows the position error and the position error bounding in the horizontal direction before FDE.

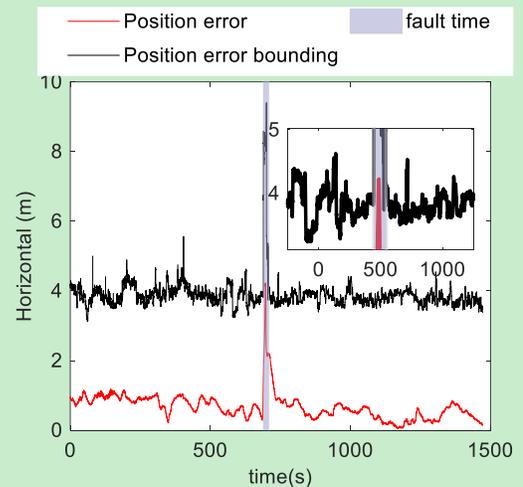

**Fig. 9.** Position error and the position error bounding in the horizontal direction before fault detection and exclusion. A step fault with amplitude of 0.15 $m/s^2$ for the accelerometer and 0.02 rad/s for the gyroscope is added to all three axes and a step fault with amplitude of 15 m is added to the pseudo-range measurement of a single satellite from 700 s to 720 s.

The results in Fig. 9 show that the position error in fault time is obviously larger than that in Fig. 6 and Fig. 8. This is due to the fact that the position error is caused by both IMU fault and GNSS fault. However, the position error bounding always envelops the position error because the position error bounding in (41) not only considers the influence of measurement noise on position error, but also the influence of faulty GNSS and IMU measurement on position error.

$\overline{GIV}$ *fault mode*

Fig. 10 shows the position error and the position error bounding in the horizontal direction after FDE when step fault with amplitude of 5 pixels is added to all the visual measurements from 180 s to 200 s and 1080 s to 1100 s.

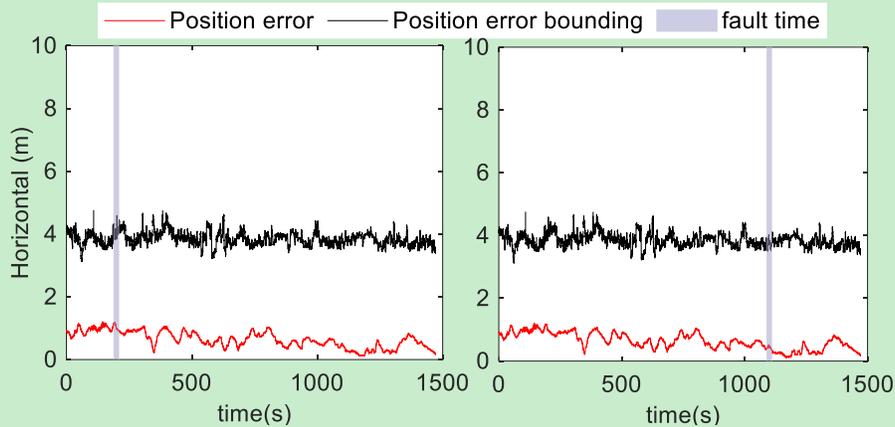

**Fig. 10**. Position error and the position error bounding in the horizontal direction after fault detection and exclusion. A step fault with amplitude of 5 pixels is added to all the visual measurements from 180 s to 200 s (left) and 1080 s to 1100 s (right)

Fig. 10 shows that the proposed position error bounding always envelops the system's position error at different fault times. Moreover, compared with $\overline{GIV}$ fault modes, the change of position error in fault time is very small. This is due to the fact that faulty visual measurements are detected and excluded immediately and GNSS/INS integration can still ensure the positioning performance of the system.

Fig. 11 shows the position error and the corresponding position error bounding in the horizontal direction before FDE.

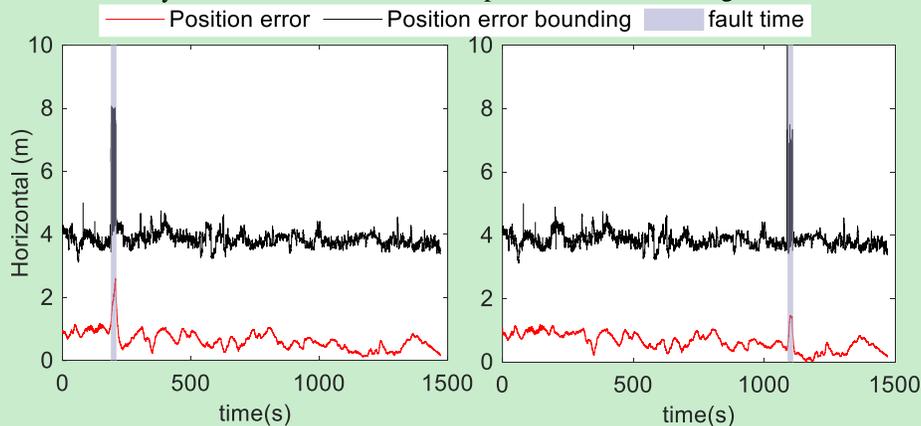

**Fig. 11.** Position error and the position error bounding in the horizontal direction before fault detection and exclusion. A step fault with amplitude of 5 pixels is added to all the visual measurements from 180 s to 200 s (left) and 1080 s to 1100 s (right)

The result in Fig. 11 indicates that the system's position error increases in fault time when faulty visual measurements are not excluded. However, the position error bounding always envelops the position error. This is due to the fact that position error bounding in (40) considers the influence of the faulty and non-faulty visual measurements on position error.

$\bar{G}IV$ *fault mode*

The previously mentioned IMU fault and visual fault are added to the IMU measurements and all the visual measurements from 700 s to 720 s, respectively. Fig. 12 shows the position error and the position error bounding in the horizontal direction before FDE.

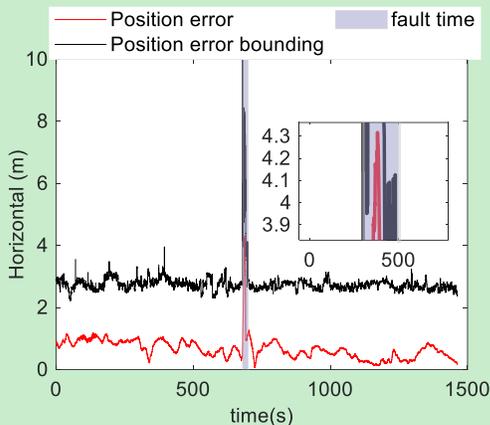

**Fig. 12.** Position error and the position error bounding in the horizontal direction before fault detection and exclusion. A step fault with amplitude of 0.15 $m/s^2$ for the accelerometer and 0.02 rad/s for the gyroscope is injected to all three axes and a step fault with amplitude of 5 pixels is added to all the visual measurement from 700 s to 720 s

The same as the position error in Fig. 9, the results in Fig. 12 show that the position error in fault time is obviously larger than that in Fig. 6 and Fig. 11. The reason is that the result is caused by both IMU fault and visual fault. Moreover, the position error bounding also envelops the position error because the position error bounding in (42) not only considers the influence of measurement noise on position error, but also the influence of faulty visual and IMU measurement on position error.

Table 1 shows the average position error, average position error bounding, and the availability of the integrity in the six fault modes after FDE, and the AL of the system is 6 m.

TABLE I
SUMMARY OF RESULTS – POTION ERROR BOUNDING AND PERCENTAGE AVAILABILITY

| Parameter | $P_{HMI,Total} = 10^{-7}/h$、AL = 6 m、$P_{MD} = 10^{-3}/h$、$P_{FA} = 10^{-5}/h$ | | | | | |
|---|---|---|---|---|---|---|
| Fault mode | $\overline{GIV}$ | $G\overline{IV}$ | $GI\overline{V}$ | $\overline{GIV}$ | $\bar{G}IV$ | $\bar{G}I\bar{V}$ |
| Average error (m) | 0.65 | 0.65 | 0.67 | 0.64 | 0.66 | 0.64 |
| Average position error bounding (m) | 5.49 | 3.50 | 3.87 | 3.85 | 2.76 | 4.54 |
| Availability (%) | 100 | 100 | 100 | 100 | 100 | 100 |

The availability in Table 1 is obtained by comparing each position error bounding with AL of 6 m. There is a negligible difference in the position error between the six fault modes. The reason is that the corresponding faults are detected and excluded in time, and the availability of integrity in six fault modes is 100%.

## V. CONCLUSION

In this contribution, we derived the specific position error bounding formula for the tightly-coupled GNSS/INS/Vision integration based on FGO. The public GVINS dataset "sport field" was used to evaluate the performance of the position error bounding, and the conclusions are summarized as follows:
(1) For the tightly-coupled GNSS/INS/Vision integration based on FGO, we derive the position error bounding formula based on the linearized equations from the residuals of GNSS pseudo-range, IMU pre-integration and visual measurements for the first time. The results demonstrate that the proposed position error bounding assures the integrity of the integrated system whether there are faults on GNSS pseudo-range, IMU pre-integration and visual measurements.
(2) For different fault modes, the position error of the system in fault time changes significantly before and after fault detection and exclusion. No matter the faults are excluded or not, the proposed position error bounding always envelop the position error.
(3) For six fault modes, there are little difference in the horizontal position errors. Additionally, the availability of their integrity is 100% by comparing each position error bounding with AL of 6 m after fault detection and exclusion.

ACKNOWLEDGMENT

This study was financially supported in part by the National Key Research and Development Program of China (Grant No. 2023YFB3907302). We would like to thank the HKUST Aerial Robotics Group for providing the field test data.

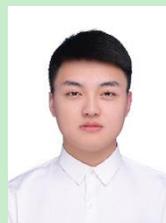

**Yunong Tian** received a Bachelor's degree from Tianjin University, Tianjin, China, in 2022. He is currently pursuing a Master's degree in Control Engineering at Beijing Institute of Technology, Beijing, China. His current research primarily focuses on GNSS/INS/Vision integration navigation and integrity monitoring.

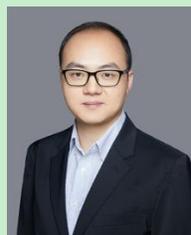

**Tuan Li** obtained his Ph.D. from the GNSS Research Center of Wuhan University in 2019, and is currently an associate research fellow with the Beijing Institute of Technology, Beijing, China. His current research focuses on precise BeiDou/GNSS positioning, GNSS/INS/Vision integration, pedestrian navigation, integrity monitoring, and multi-sensor fusion for robotics.

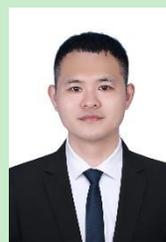

**Haitao Jiang** received an M.S. degree from Taiyuan University of Technology, Taiyuan, China, in 2019. He is currently working toward a Ph. D. degree with the School of Electronics and Information Engineering, Beihang University. His current research mainly involves multi-source fusion navigation and integrity monitoring, which includes GNSS, inertial and visual integrated navigation.

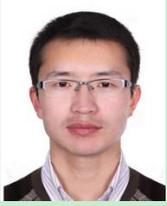
Zhipeng Wang received his B.S. in Communication Engineering from Northwestern Polytechnical University (NWPU) in 2006, and Ph.D. in Traffic Information Engineering from Beihang Univeristy in 2013. He is a professor of the Department of Electronic Information Engineering, Beihang Univeristy, China. His research focuses on high precision aviation navigation Minimum Operation Network (MON), aviation navigation integrity monitoring, GNSS aviation application test verification and international standardization.

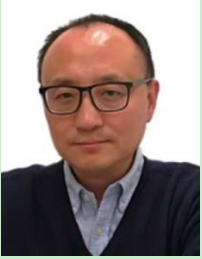
Chuang Shi received the B.Eng. degree in geodesy and the Ph.D. from Wuhan University, Wuhan, China, in 1991 and 1998, respectively. He has worked as an expert in the field of earth observation and navigation for the 863 Program, and the Deputy Chief Designer of the National BeiDou Ground-Based Augmentation System. He is currently a professor and Ph.D. supervisor of the School of Electronic and Information Engineering, Beihang University. His research interests cover high-precision GNSS theories and applications, including network adjustment, precise orbit determination of GNSS satellites and LEOs and real-time precise point positioning (PPP).